\documentclass[conference]{IEEEtran}
\IEEEoverridecommandlockouts
\usepackage{cite}
\usepackage{amsmath,amssymb,amsfonts}
\usepackage{algorithmic}
\usepackage{graphicx}
\usepackage{textcomp}
\usepackage{xcolor}
\def\BibTeX{{\rm B\kern-.05em{\sc i\kern-.025em b}\kern-.08em
    T\kern-.1667em\lower.7ex\hbox{E}\kern-.125emX}}
\newcommand{\etal}{{\em et al.}}
\begin{document}

\title{Application of deep learning to camera trap data for ecologists in planning / engineering -- Can captivity imagery train a model which generalises to the wild?\\

\thanks{Research funded by Ove Arup and Partners}
}

\author{\IEEEauthorblockN{Ryan Curry}
\IEEEauthorblockA{\textit{Consulting North} \\
\textit{Ove Arup and Partners}\\
Newcastle, UK \\
ryan.curry@arup.com}
\and
\IEEEauthorblockN{Cameron Trotter}
\IEEEauthorblockA{\textit{School of Engineering} \\
\textit{Newcastle University}\\
Newcastle, UK \\
c.trotter2@newcastle.ac.uk}
\and
\IEEEauthorblockN{A. Stephen McGough}
\IEEEauthorblockA{\textit{School of Computing} \\
\textit{Newcastle University}\\
Newcastle, UK \\
stephen.mcgough@newcastle.ac.uk}
}

\maketitle

\begin{abstract}
Understanding the abundance of a species is the first step towards understanding both its long-term sustainability and the impact that we may be having upon it. Ecologists use camera traps to remotely survey for the presence of specific animal species. Previous studies have shown that deep learning models can be trained to automatically detect and classify animals within camera trap imagery with high levels of confidence. However, the ability to train these models is reliant upon having enough high-quality training data. What happens when the animal is rare or the data sets are non-existent? This research proposes an approach of using images of rare animals in captivity (we focus on the Scottish wildcat) to generate the training dataset. We explore the challenges associated with generalising a model trained on captivity data when applied to data collected in the wild. The research is contextualised by the needs of ecologists in planning / engineering. Following precedents from other research, this project establishes an ensemble system of object detection, image segmentation and image classification models which are then tested using different image manipulation and class structuring techniques to encourage model generalisation. The research concludes, in the context of Scottish wildcat, that models trained on captivity imagery cannot be generalised to wild camera trap imagery using existing techniques. However, final model performances based on a two-class model (\texttt{Wildcat} vs \texttt{Not Wildcat}) achieved an overall accuracy score of 81.6\% and \texttt{Wildcat} accuracy score of 54.8\% on a test set in which only 1\% of images contained a wildcat. This suggests using captivity images is feasible with further research. This is the first research which attempts to generate a training set based on captivity data and the first to explore the development of such models in the context of ecologists in planning / engineering.
\end{abstract}


\section{Introduction}
It is almost now without question that humans are having a major impact on other species living on the planet~\cite{ipbes_2019_5657041}, with many species at significant risk of going extinct within the next ten years. Identifying the abundance of a particular species in an area allows us to estimate its sustainability as well as assessing the impact humans are having upon that species. This becomes ever-more difficult as a species becomes more rare, yet these are the species we need to monitor the most.

In the context of planning / engineering, ecologists use camera traps to remotely survey for the presence of animal species which have elevated protection levels under national and international law. Camera traps are placed in an area of interest for a period of time and are motion activated (take a picture / record video whenever motion is detected in front of the camera)~\cite{tabak_machine_2019}. This work focuses on camera trap surveys undertaken by ecologists in Scotland to monitor the abundance of the Scottish wildcat. However, we see the work here being easily transferable to other species.

The Scottish wildcat (\textit{Felis silvestris}) is the last member of the \textit{Felidae} (cat) family living wild in Britain~\cite{fredriksen_wildcats_2016}. Estimates in 2005 suggested there were as few as four-hundred left in the wild, with this number believed to have fallen further since. The reduction in numbers is largely due to interbreeding with domestic cats, resulting in an increasing population of ``hybrid'' wildcat and feral cats, thus leading to a dwindling number of pure wildcats~\cite{kilshaw_detecting_2015}. Under the UK’s Wildlife and Countryside Act 1981, pure wildcats have elevated levels of protection and engineering projects must mitigate their impact upon them if they are detected within the project’s proximity~\cite{kitchener_diagnosis_2005,noauthor_wildlife_1981}. Hybrid wildcat / feral cats hold no such protection. 

Whilst camera traps are an effective, non-invasive way of surveying for the presence of wildcat, they suffer from a high number of ``false'' activations caused by motion of background vegetation (e.g., a tree on a windy day), non-target species or other nondescript events~\cite{beery_efficient_2019}. This results in the cameras collecting potentially thousands of images, a significant proportion of which contains no valuable information and requires filtering out. Informal experiments conducted for this paper found that ecologists can manually review approximately 75 images in 3 minutes. Camera trap surveys can result in datasets ranging from tens of thousands to millions of images \cite{norouzzadeh_deep_2021}. Thus, the time required to review all images varies from a full working day to several months. Furthermore, human fatigue and general human error increasingly result in misclassifications. Making automation of the process very attractive.

The opportunity to automate camera trap data processing has seen a significant amount of research over the years, progressively focusing on statistical methods, traditional computer vision techniques and deep learning techniques~\cite{christin_applications_2019}. A large proportion of existing research was found to focus on extremely large datasets, such as those in the publicly available datastore LILA BC~\cite{noauthor_lila_nodate}. In that context, developing models capable of classifying animals in imagery with accuracies over 90\% are considered as normal~\cite{schneider_past_2019}. Whilst this is a great outcome, it poses an issue for ecologists who are working with real-world data -- such as those in the planning / engineering realm. In such realms, if it cannot be said with an extremely high degree of confidence that an automated method can correctly identify \textit{every} image which \textit{possibly} contains a species of interest (or at least within the margin of error of manual review), legally and morally ecologists would still be obligated to check every image manually.  

None of the reviewed public dataset repositories (LILA BC~\cite{noauthor_lila_nodate} or Google Open Images~\cite{google_google_nodate}) contained camera trap imagery of Scottish wildcat. Though there are many pictures of domestic cats which could easily be confused for the Scottish wildcat. Due to the rare nature of wildcat, attempting to develop a dataset of such images comprehensive and large enough to train a model would require an extremely large camera trap survey over an extended period of time. However, Scottish wildcat are held in captivity in numerous locations in the UK, making the development of a camera trap image dataset more feasible. This therefore raises the question of can a model trained on images of captive animals be used for training an identification tool which is applied to camera trap surveys in the wild? If so, this significantly raises the feasibility of developing automated approaches to classify such rare species. 

This paper seeks to advance research in the automation of processing camera trap imagery by exploring if a model trained on images of captive wildcat can be generalised to the wild. The research is undertaken in the context of ecologists in the realm of planning / engineering and their need to identify target species with a high level of confidence.

Section \ref{chap:relwork} details a review of related work, Section \ref{chap:methodology} describes the methodology followed to undertake the research, Section \ref{chap:results} provides the research results, and Section \ref{chap:conc} draws conclusions on the research to date and provides suggestions for further study.

\section{Related Work} \label{chap:relwork}
\subsection{Scottish Wildcat}
\begin{figure}[htbp]
    \centering
    \centerline{\includegraphics[scale=0.25]{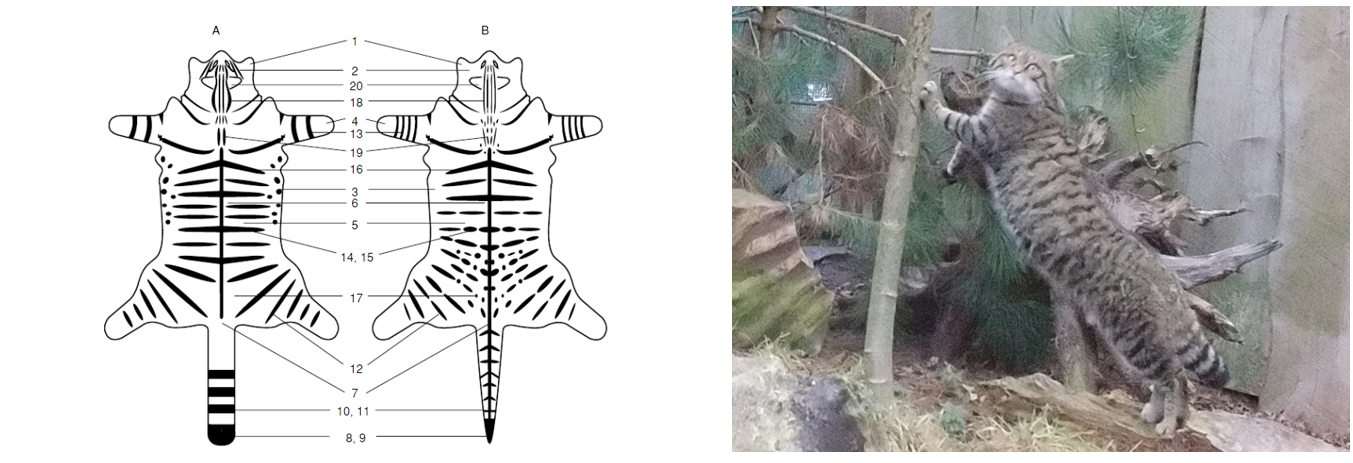}}
    \caption{(Left) Pelage scoring diagram devised in \cite{kitchener_diagnosis_2005} to distinguish pure wildcat (left) from hybridised wildcats (right). (Right) example of pure wildcat in captivity displaying pelage traits.}    
    \label{fig:lit_wildcat}
\end{figure}

Although the wildcat is similar in appearance to other members of the cat family it is possible to distinguish between them using fine-grained image classification~\cite{trotter2020ndd20,8634790}. The opportunity to achieve this fine-grained image classification system for Scottish wildcat is made possible by the fact wildcat are visually distinct, with light brown fur and black markings~\cite{kitchener_diagnosis_2005}. This work by Kirchener \etal stands as a seminal piece of research for ecologists, describing the characteristics of a wildcat’s pelage (i.e., it’s fur) which allows one to identify a pure wildcat and distinguish it from a hybridised wildcat. Figure \ref{fig:lit_wildcat} (left) shows the location and quantity of stripes / spots expected. Figure \ref{fig:lit_wildcat} (right) is a picture of a pure wildcat in captivity.

Recent genetic research has shown that pelage scoring may no longer be reliable in determining the purity of a wildcat due to the depth of interbreeding now present~\cite{senn_distinguishing_2019}. However, as DNA sampling is impractical on a scale comparable with camera trapping, visual identification by camera trapping is still a widely accepted form of wildcat identification for ecologists in planning / engineering. This, in turn, validates the proposal to develop a model to automate sifting of imagery for potential wildcat.

\subsection{Object Detection and Image Classification}
There has been interest in applying deep learning techniques to processing camera trap imagery for several years. The work by Christin \etal~\cite{christin_applications_2019} details a review of 87 papers using deep learning for ecological identification, 59 of which focus upon imagery. It found the first mention of deep learning in an ecological context in the year 2000, however, most research using deep learning has been published since 2015.

Deep learning approaches initially focused on developing models on entire camera trap images. Some developed a single image classification model and others developed a detection classifier (contains or doesn’t contain an animal) which was run in tandem with the species classifier as an ensemble model~\cite{beery_recognition_2018,christin_applications_2019,nguyen_animal_2017,tabak_improving_2020}. This research predominately focused on training and testing on single, large datasets, such as Snapshot Serengeti~\cite{swanson_snapshot_2015}. Whilst performance was shown to be reasonable in sample, there was a notable drop in performance in research that performed out of sample testing~\cite{beery_recognition_2018,norouzzadeh_deep_2021}. In particular, Tabak \etal~\cite{tabak_improving_2020} showed a performance accuracy of 96.8\% for their species classifier with in sample testing, however, performance dropped to 36\%, 50.7\%, 56.2\% and 91\% across the four out of sample datasets tested. We concur with the findings of this works demonstrating tha the generalisation issue was also particularly acute in the research undertaken, with captivity imagery having limited background variability and the artificial nature of the enclosure unlike the background imagery found in the wild.

In a bid to improve model performance (and indirectly overcome the generalisation problem), research progressed to training object detection models to first localise an animal within an image, then classify said animal~\cite{beery_recognition_2018,shepley_automated_2021,tabak_machine_2019}. This approach appeared to improve performance over whole image classifiers, with Tabak \etal~\cite{tabak_machine_2019} showing top-1 classification accuracies (the predicted label with the highest confidence being correct) improving from 79.18\% to 91.86\% for full image and object detection bounding box cropped images respectively -- testing was conducted on the same dataset. However, Tabak \etal also detailed the failure of object detection models to generalise, showing mAP scores falling from 77\% to 71\% when tested in sample compared to out of sample.

In an attempt to overcome issues around generalisation, various approaches have been trialled. Beery \etal~\cite{beery_efficient_2019} details an object detection model (MegaDetector) trained on 4.8 million images to simply detect the presence of animals as a single class. These images came from a wide range of contexts and species, with the goal to provide a model which generalises as widely as possible. The mAP scores reported by the paper ranged from 88.5\% to 98.8\% dependant upon the dataset tested. With independent research applying the model reporting accuracies of 91.71\%~\cite{norouzzadeh_deep_2021}. Another approach to enhance generalisation is to infuse camera trap data with imagery from other sources, such as Flickr\footnote{https://www.flickr.com} and iNaturalist\footnote{https://www.inaturalist.org}. Shepley \etal~\cite{shepley_automated_2021} details this work, showing that infusing camera trap data with up to 15\% of imagery from other sources shows mAP performance increases of 3.66\% to 18.20\%, although improvements plateaued or decreased when infusion proportions increased further. 

In addition to published research papers, an additional avenue to understand current thinking in the field is the iWildcam competition~\cite{beery_iwildcam_2021}. iWildcam is an annual competition, a part of the workshop on Fine-Grained Visual Categorization (FGVC) at the Conference on Computer Vision and Pattern Recognition (CVPR), which seeks to incrementally advance research into camera trap image processing by focusing on specific areas each year\footnote{https://github.com/visipedia/iwildcam\_comp}. A review of 2021 submissions found most followed a deep learning ensemble approach, classifying full images and cropped images, relying upon MegaDetector to provide the image cropping proposals. This supports previous work which follows an ensemble approach.

\subsection{Image Segmentation}
An approach suggested by Shepley \etal~\cite{shepley_automated_2021} for additional research, as well as provided as an option in the 2021 iWildcam competition, was the use of image segmentation to overcome issues around background imagery causing generalisation issues. Image segmentation is the process in which a statistical or deep learning model attempts to separate an image in the foreground from background imagery. Classical computer vision methods include the GrabCut algorithm~\cite{rother_grabcut_2004}, which employs iterated graph cuts to segment an object in a bounding box from the background. Deep learning approaches which attempt to create a ``mask'' over the object in question, informed by the location of a bounding box include the Mask-RCNN architecture~\cite{he_mask_2018} (which detects the objects itself and masks it) and DeepMAC~\cite{birodkar_surprising_2021}, a CenterNet~\cite{duan_centernet_2019-1} architecture model (pretrained on the COCO~\cite{lin_microsoft_2015-1} classes) which takes bounding boxes as an input and has been shown to perform well for generalised classes. All three methods are examples of instance segmentation. Application to the captivity images could prove beneficial should model generalisation be an issue. 

\section{Methodology} \label{chap:methodology}
\subsection{Overview}
The development of the research methodology was primarily informed by three key requirements. Firstly, the developed system needs to be able to identify as many images as an ecologist manually could which \textit{possibly} contain wildcat. This is to enable the ecologists to concentrate their work in documenting the locations and numbers of wildcat. Secondly, the system should classify any image which \textit{possibly} contains any other animal, as containing an animal. This includes images in which the type of animal is visually identifiable and images in which something indicates an animal (e.g. a non-descript blur), but the exact species is not identifiable. As the range of animals possibly captured is not known, the classification needs to be general. Finally, any empty background image should be identified as being empty.

The key differentiator in this research was that the imagery of wildcats for model training would come exclusively from a captive enclosure. Whilst generalisation is a concern in any classification system, the limitations associated with image capture in captivity are particularly acute. The other main differentiators are the logic used for classification and the scope of the model classes. In terms of the classification logic, reviewed research attempts to achieve the highest possible accuracy evenly across all classes. For this research, there is a desire for a classification hierarchy, in which lower overall model accuracy would be acceptable if the number of false negatives in the wildcat class can be reduced. With regards to the scope of the classes, the approach differs from the main body of research which attempts to either classify generally (is there an animal or not) or classify which species is present. A blend of the two has not been previously seen, to the best of the authors knowledge. This helps achieve the overall requirements of this work.

From the findings of the literature review, it was decided to explore the development of a deep learning ensemble classification system. The two drivers for this were that an ensemble system can combine the benefits of different types of models to improve overall system performance and could include redundancies to retain images which may be falsely discarded.

Section \ref{chap:ensemble} explains the system the research would attempt to deliver, Section \ref{chap:thedata} describes the data available to undertake the research, and Section \ref{chap:researchapp} describes the methodology of developing the system using the data. 

\begin{figure*}[t]
    \centerline{\includegraphics[scale=0.45]{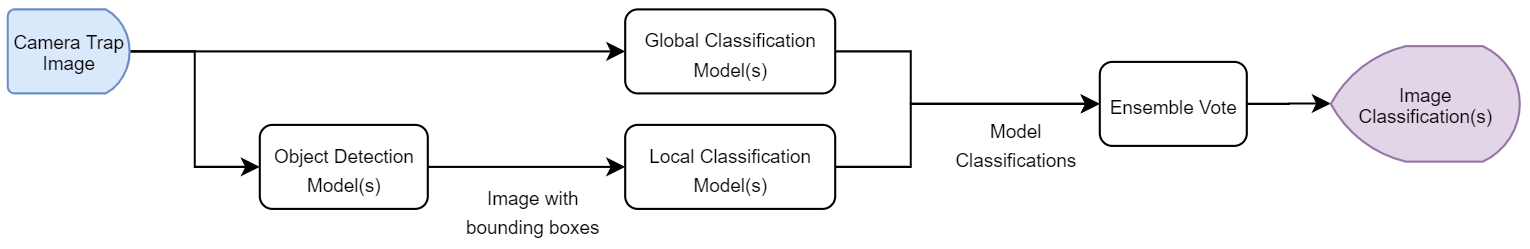}}
    \caption{High level workflow of proposed model architecture.}    
    \label{fig:meth_wf1}
\end{figure*}
\begin{figure*}[t]
    \centerline{\includegraphics[scale=0.4]{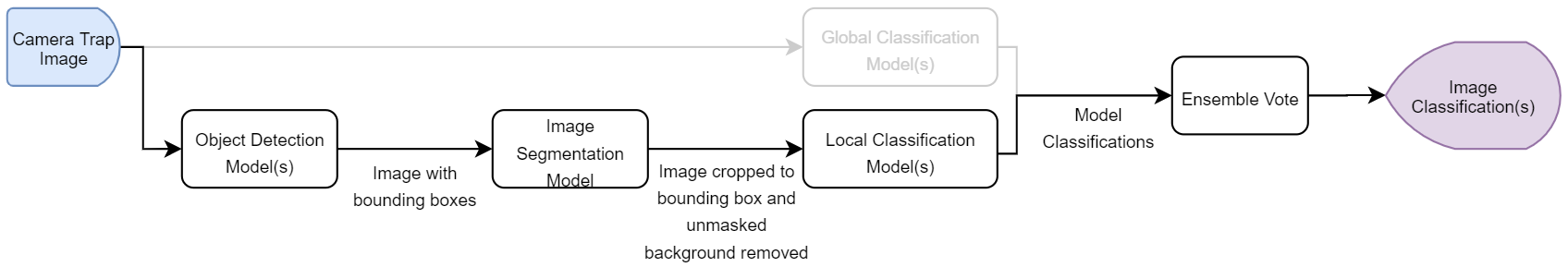}}
    \caption{Local classification model workflow with image segmentation}    
    \label{fig:meth_wf2}
\vskip -10pt
\end{figure*}
\subsection{Ensemble System Structure Description} \label{chap:ensemble}
The architecture of system researched in the project attempts to classify images both on a local and global level. The global model(s) attempt to classify an image in its entirety, whilst the local model(s) attempt to classify an animal found within a specific part of an image. The local models are fed by the bounding box(es) outputs from object detection model(s). A vote between the results of the models allows the whole system to decide on the resulting classification(s) of each camera trap image. The focus of the research therefore explores which combination of model architectures should be combined to create the best overarching system architecture. Figure \ref{fig:meth_wf1} displays a high-level workflow of the proposed system, showing how the different models link together to produce the overall system.

Local classification models were considered due to the literature review showing images cropped to the bounding box of an animal are more likely to train a model which generalises to new locations than a model trained on full camera trap images. The inclusion of a local model therefore predicated the requirement for an object detection model to provide suitable region proposals to classify. The global model aspect of the system was included to ensure every image was seen by a classifier. Whilst whole image classifiers were shown to be less reliable in the reviewed research~\cite{norouzzadeh_deep_2021,beery_recognition_2018}, the reliance on an object detection model prior to the local classification would inherently mean some images would be falsely disgarded prior to the local classification model having sight of them. Combining the strengths of all three models was seen as the best starting point for the system.

An additional component researched was the inclusion of an image segmentation model as part of the local model pipeline. This attempts to mask out the background around any potentially identified animal from the object detection model. This component was included due to the background being believed (as a result of priort research) to be a key aspect which inhibits models from generalising. Figure \ref{fig:meth_wf2} shows how the instance segmentation model fits into the ensemble system.

\subsection{The Data} \label{chap:thedata}
Three image datasets were curated for use during the project. Their collation and initial processing is described below, including how bounding boxes were created for training the local model aspect of the ensemble system. 
\begin{figure*}[t]
    \centerline{\includegraphics[scale=0.55]{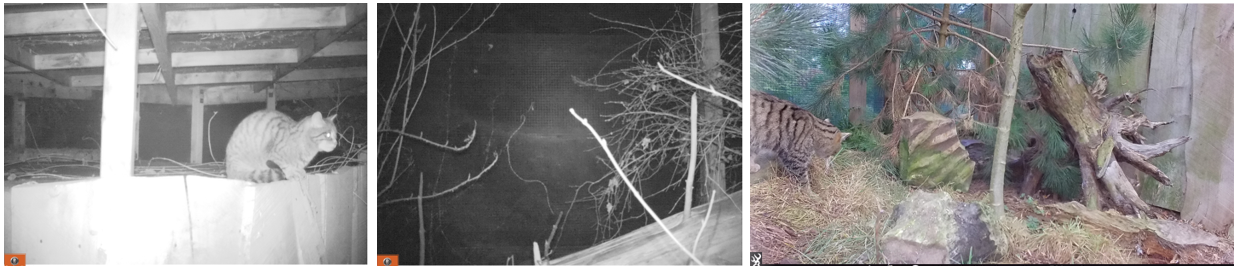}}
    \caption{Captivity camera trap perspectives (Camera 1, 2 and 3 respectively)}
    \label{fig:meth_traps}
\end{figure*}

\subsubsection{Captivity Dataset} \label{chap:wildcatdata}
To generate a dataset to undertake the project, three camera traps were placed in a captive wildcat enclosure at Calderglen Country Park, East Kilbride, Scotland. Over a seven day period, the cameras collected 3,890 images from three perspectives. Figure \ref{fig:meth_traps} displays examples of the perspective captured from each camera.

The images were manually annotated using the python package labelimg \cite{tzutalin_labelimg_2021}, with bounding boxes created for each category identified. Two sets of labels were developed to enable more nuanced testing to be undertaken. Set 1 categorised any object potentially identifiable as a wildcat as either \texttt{Wildcat (daytime colour)} or \texttt{Wildcat (night-time IR)} (IR - infrared). Any non-descript motion blurs were classified as \texttt{Animal Unknown} and any empty images were subsequently classified as \texttt{Background}. This set aligned with the stated aim of classifying any \textit{possible} wildcat as being wildcat, then any \textit{possible} animal as being an animal. Set 2 used the same labels but changed the definition of how the \texttt{Wildcat} classes were defined to a more strict interpretation. This resulted in some images moving from the wildcat classes into the \texttt{Animal Unknown} class. Set 2 sought to counter the bias of knowing that all movement in the enclosure was a wildcat, therefore creating a \texttt{Wildcat} class which showed more strongly defined features. Images referred to as being \texttt{Wildcat} should be considered as \texttt{Wildcat (daytime colour)} and \texttt{Wildcat (night-time IR)} images combined unless stated otherwise.

To provide \texttt{Background} imagery to the local models, bounding boxes of random size were generated for images without wildcat and from areas outside existing bounding boxes for images with \texttt{Wildcat} or \texttt{Animal Unknown}. The Captivity dataset was used to populate this class as the image had been confirmed to be empty (as a product of the wildcat classification) and using the same imagery for \texttt{Background} and \texttt{Wildcat} could potentially assist the models learn what distinguishes the presence of an animal.

A feature of this dataset is that the camera traps take images in \textit{bursts} -- through configuration of the camera. Camera 1 and 2 captured three images per activation, with Camera 3 capturing eight images per activation. If an animal remained relatively stationary during the burst, this generated a number of very similar images. In a similar vein, the habits of the wildcat (such as pacing round the perimeter of the enclosure) can also lead to a series of very similar images across different bursts, as the camera is activated at the same time. Consideration of this was required during development of the ensemble system methodology to ensure training sets weren't cross-contaminated.

\subsubsection{General Animals (Not Wildcat)}
A general \texttt{Animal Other} dataset was created using the iNaturalist 2018 training set~\cite{van_horn_inaturalist_2018}. Two-thousand images were randomly sampled from the dataset for use in the modelling. The number selected aligned with the proportion of the \texttt{Wildcat} and \texttt{Animal Unknown} images collected. Time was not available to manually classify the imagery in this dataset, so the MegaDetector model was applied and the highest confidence bounding box adopted. A review of the dataset found that there was some minor pollution, such as sample images containing animal prints or reference objects (rulers, fingers, Swiss army knives etc.) rather than animals themselves. There were sixty-five animal classes present in the dataset, spread across birds and mammals. While the dataset contained animals from different perspectives / proportions of the image, the dataset was predominately photographs with camera trap data being in the minority. None of the sixty-five classes contained wildcat or domestic cats.

\subsubsection{Wild Test Set}
Data \textit{from the wild} comes from a camera trap survey undertaken in Scotland\footnote{Details around the location and positioning of wild camera traps cannot be identified due to interpretation of legislation around the protection of wildcat.}. For this research project, unlabelled images from twelve perspectives were available for use. Due to this data being from the wild, bounding box image classification was required to be undertaken by ecologists. Four-thousand of the images (randomly sampled from the full set) were labeled. Labelling was undertaken using labelimg, with the environment set up and coaching of how to use the software provided as part of the research. From the four-thousand images, 42 images were labelled as potentially containing \texttt{Wildcat}, 681 labelled as \texttt{Animal Other} and 295 labelled as \texttt{Animal Unknown}. The \texttt{Wildcat} class therefore represents only 1\% of the overall Wild test set.

\subsection{Research Approach} \label{chap:researchapp}
Research development of the ensemble system was separated into two phases. Phase 1 (P1) explored the different architectures that could be used to construct the elements of the ensemble system. Phase 2 (P2) makes use of the point labelling from the Wild test set as it became available, which enabled systems explored in Phase 1 to be applied \textit{to the wild}. Experiments in Phase 2 were aimed at improving the model generalisation and were informed by sequentially applying knowledge gained from previous tests. The aspects explored have been grouped into the \textit{Tests} detailed below. %\todo{CT Comment: Personally I would remove reference to work that was abandoned and use it as future work}

\subsubsection{Test 1 (P1) - Object detection model selection} Initially, the potential to train an object detection model using an existing architecture (either from scratch or via transfer learning) was explored. However, this was discontinued due to project time constraints. Instead, the performance of existing object detection models were assessed, on the principle that their raw pre-trained performance may be enough to inform development of subsequent elements of the system. The approach used the models as general region proposal models in the ensemble system, ignoring the classification produced and passing all bounding box predictions to the local model. This would provide the local models the highest chance of being exposed to a potential animal in each camera trap image. Object detection models trained to two sets of data were considered – MegaDetector~\cite{beery_efficient_2019}, a Faster-RCNN~\cite{NIPS2015_14bfa6bb} ResNet model trained to have a general animal class (along with a person and vehicle class), and 13 models architectures trained on the COCO dataset (which among other categories contains a \texttt{cat} class). The COCO models chosen represented a breadth of both model types and stated performance (against COCO). 

\subsubsection{Test 2 (P1) - Image cross-contamination} Dataset cross-contamination was a concern due to the issues around stationary animals in the same burst or repetitive movement across multiple bursts causing similar imagery to be generated. However, there is no objective way to define how similar is \textit{too} similar with regards to model cross-contamination. To explore this, models were trained using local and global model imagery from the Captivity dataset, with the data separated in four different ways. The \textit{burst-based} approach saw images separated by the burst in which they were taken, such that all images taken in the same burst would be placed into either the training set, validation set or test set (based on a 70\%-20\%-10\% split). The \textit{camera hold-out} approach saw all images from a single camera placed into the test set, with images from the remaining cameras placed into training and validation. Tests were undertaken holding out images from each of the three cameras respectively. Performance was measured via overall model accuracy. The outcome of this test would inform the structure of training sets for subsequent tests.

\subsubsection{Test 3 (P1) - Local and global model architecture} The approach for researching the classification models was to test which model architectures would perform best and how many models working together would perform best. Eight architectures were considered: DenseNet201 \cite{huang_densely_2018}, EfficientNetB7 \cite{tan_efficientnet_2020}, InceptionV3 \cite{szegedy_rethinking_2015}, MobileNet \cite{howard_mobilenets_2017}, MobileNet-V3Large \cite{howard_searching_2019}, ResNet152V2 \cite{he_deep_2015}, VGG19 \cite{simonyan_very_2015} and Xception \cite{chollet_xception_2017}. These architectures were selected as they represented a reasonable range both in terms of model size and between proven vs new approaches. The number of models to possibly combine in the ensemble was limited to three local models and three global models. This was primarily due to practicality concerns, in that an unwieldy system with many models would not have much real world practicality. The decision on the final number to combine would be taken as part of researching the ensemble voting system. Training and testing was undertaken using the Captive test set, with testing on the Wild test set for each model architecture also retrospectively undertaken.

All models implemented the pre-trained weights from the ImageNet~\cite{deng_imagenet_2009} dataset, with transfer learning applied to a new dense layer of 1024 units (using ReLU activation) and a final softmax output layer. Experiments were undertaken using the Adam optimiser~\cite{kingma2017adam}. Experimenting with other combinations of final model layers and transfer learning deeper into the model was considered. However, the hyper local nature of the training set meant that training to any degree was highly likely to be susceptible to overfitting and adjustments of the dataset were considered to have a greater effect on the model than tweaking the final layers of the model itself. 

\subsubsection{Test 4 (P1) - Ensemble voting system} Two approaches for developing a voting system (based on the predictions of the local and global models) were explored: A conservative approach which favoured the votes of models predicting a higher priority category (i.e. \texttt{Wildcat} \textgreater \texttt{ Animal Other} \textgreater \texttt{ Animal Unknown}  \textgreater \texttt{ Background}) and an accuracy approach which sought to take the majority vote as its classification.

An aspect of researching the voting system was also to decide the number and type of models to be used from the local and global model experimenting. For the maximum of the three models from the local and global sides of the system, performance accuracies for all combinations of architectures were considered – including three of the same model (thereby checking if one architecture itself outperformed any other combination) and two of one architecture and one of another (thereby checking if a weighted priority vote of one system in combination with another was more effective). 

\subsubsection{Test 5 (P1) - Segmentation application} The successful use of a segmentation model in the system would be reliant upon the segmentation model returning a well segmented image (mask successfully removing background \textit{around} the animal). This test explored two approaches – GrabCut and DeepMAC. Both were implemented `as is', with no additional training other than to pass the relevant bounding boxes to enable the masking processes to take place. The performance of these approaches were reviewed manually as no ground truth masks existed to systematically assess the predictions.

\subsubsection{Test 6 (P2) - Dataset class adjustment - Captivity Dataset \textit{Set} and Wildcat class separated} Test 6 explored the impact on model performance by adjusting the model classes. Test 6a explored the effect of moving from Captivity Dataset Set 1 to Captivity Dataset Set 2. This tested whether a more generalised class increased the likelihood of a model predicting that category, or confused it by making the class less distinct. See Section \ref{chap:wildcatdata} for more details on Set 1 and 2. Test 6b explored the effect of separating the daytime colour and night-time IR imagery contained in the \texttt{Wildcat} class into individual classes to create more distinct classes. Both tests together would provide insight on the best way to structure the model classes.

\subsubsection{Test 7 (P2) - Segmented image classification} This explored training the local classification models on segmented images created by the system selected in Test 4. Training was undertaken using segmented images created with the relevant ground truth bounding boxes for the Captivity and General Animals dataset. Inference was on the Wild test set using segmented images created with bounding boxes predicted from the object detection system(s) selected in Test 1.   

\begin{table*}[tbp]
\caption{Effect on performance of models depending on dataset (HO - Hold Out)}
\begin{center}
\begin{tabular}{|c|c|c|c|c|}
\hline
\textbf{Model}&\textbf{Burst-based}&\textbf{Camera 1 HO}&\textbf{Camera 2 HO}&\textbf{Camera 3 HO} \\
\hline
\textbf{Local} &91.4\%&92.9\%&71.0\%&91.4\% \\
\hline
\textbf{Global} &84.6\%&84.3\%&44.7\%&64.1\% \\
\hline
\end{tabular}
\label{tbl:2}
\end{center}
\end{table*}

\subsubsection{Test 8 (P2) - Two class model} A final experiment undertaken was a review of the assumptions made thus far around whether using a hierarchical model class structure leads to a more robust system (e.g. a \texttt{Wildcat} would be classified as \texttt{Animal Other} rather than being discard as \texttt{Background}) or if creating a two class system (\texttt{Wildcat} and \texttt{Not Wildcat}) would perform better overall. Again, the best performing models would be selected to test.

\section{Results} \label{chap:results}
This section explores the results of the experiments described in Section  \ref{chap:methodology}. The models which performed best against the validation set were selected unless otherwise stated.

\subsection{Phase 1 Test Results}
\subsubsection{Test 1 (P1): Object detection model selection}
From undertaking the test, the MegaDetector model was found to significantly out-perform the unadjusted raw COCO-trained model architectures. As a result, MegaDector was adopted for use in developing the ensemble system and undertaking the following tests. 

%\todo{I can't see any discussion of Test1 or test 5 - there should be something even if it's just a sentence to say the results weren't interesting.}
\subsubsection{Test 2 (P1): Image Cross-contamination}
The cross-contamination test was undertaken using the MobileNet architecture for both the local and global models. Table \ref{tbl:2} displays the accuracy results of the tests. For the local model, accuracies for the burst-based, Camera 1 Hold Out (HO) and Camera 3 HO models are similar, with a notable 20\% drop for Camera 2 HO. The effect is more stark for the global models, whereby although the performance of the burst-based and Camera 1 HO are similar, Camera 3 HO experiences a performance drop of 20\% and Camera 2 HO of 40\%. In the local model context, it is believed this performance drop is actually as a result of bad camera placement, rather than an issue with the model itself. The example in Figure \ref{fig:meth_traps} shows the camera located behind a large log with branches placed directly in front of the lens. This results in the majority of images taken being of quite a low quality as the camera over-exposes the log and branches. This is compounded by the movement of the wildcat in the location, which is to walk behind the log. This results in the crops from this camera being starkly different to the other two cameras, which explains why the model fails to generalise well. Similar results are also observed for the global model, with the deteriorated performance in Camera 2 HO and Camera 3 HO likely due to generalisation issues relating unseen background imagery.

The results show that using a burst-based approach appears to allow for the benefits gained of broadening the information provided to the model without inadvertently resulting in artificially inflated accuracies. Therefore, the approach of splitting the sets by burst was adopted. 

\begin{figure}[htbp]
    \centering
    \includegraphics[scale=0.41]{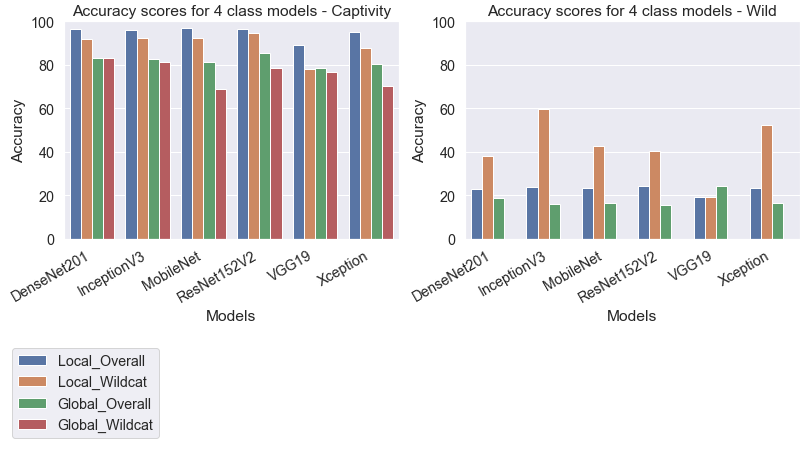}
    \caption{Accuracy performance using Captive test set (left) and Wild test set (right)}    
    \label{fig:results_arch2}
\end{figure}

\subsubsection{Test 3 (P1): Local and Global Model Architecture}
Using the burst-based approach, the eight selected model architectures were used to train models on both the local and global scales. However, initial experimentation found that EfficientNetB7 and MobileNetV3Large performed poorly and thus they were not considered further. The models were trained on four classes: three from the Captivity set (\texttt{Wildcat}, \texttt{Animal Unknown} and \texttt{Background}) and the fourth from the General Animals dataset (\texttt{Animal Other}). Figure \ref{fig:results_arch2} displays the performance of the Captivity test set and the Wild test set respectively. The overall model accuracy and that of the \texttt{Wildcat} class is reported, as this is the class which is of primary interest.

Figure \ref{fig:results_arch2} shows the performance of the architectures on the Captivity set to be good, with the local models scoring highly, in the 90-95\% range, and global models in the 70-85\% range. The gap in performance between the local and global models was expected, both as a result of insight from Test 2 and alignment with previous research. When the Captive test set results are compared to the Wild test set, a significant deterioration is observed. This appears to confirm concerns around generalisation when testing a model trained on captive imagery to the wild. Exploration of the confusion matrices found a strong preference for the \texttt{Animal Other} class across all ground truth classes. Reviewing the imagery, it is believed that the strong variety both in animal species and backgrounds in the \texttt{Animal Other} imagery creates a powerfully diverse class. Whilst having a well generalised \texttt{Animal Other} class is desirable to identify animals in the Wild which the model hasn't previously seen, the matrices suggest the \texttt{Wildcat} class isn’t distinctive enough to stop \texttt{Wildcat} images also being drawn to the \texttt{Animal Other} class. A similar trend is believed to be occurring with the \texttt{Background} class, in that the \texttt{Animal Other} class has been trained on a more noisy, diverse range of images.

Due to the performance issues experienced during this test, it was decided to focus efforts upon the local model aspect of the system as this displayed better performance overall. As such, no further development of the global model was undertaken.

\begin{table*}[tbp]
\caption{Wild test set accuracy scores for Tests 3, 6-8}
\begin{center}
\begin{tabular}{|c|c|c|c|c|}
\hline
\textbf{Test ID}&\textbf{Description}&\textbf{No of Classes}&\textbf{Accuracy – Overall}&\textbf{Accuracy – Wildcat} \\
\hline
\textbf{3}&Local model architecture&4&22.7\%&40.5\% \\
\hline
\textbf{6a}&Dataset Class Adjustment - Captivity Dataset Set&4&22.4\%&45.2\% \\
\hline
\textbf{6b}&Dataset Class Adjustment - Wildcat class separated&5&25.1\%&28.6\% \\
\hline
\textbf{7}&Segmented image classification&4&12.9\%&47.6\% \\
\hline
\textbf{8a}&Two Class Model (Segmented)&2&81.6\%&54.8\% \\
\hline
\textbf{8b}&Two Class Model (Local Crop)&2&{\bf 98.7\%}&3.1\% \\
\hline
\textbf{8c}&Two Class Model (Animal Unknown Removed – Segmented)&2&64.5\%&{\bf 76.2\%} \\
\hline
\end{tabular}
\label{tbl:6}
\end{center}
\end{table*}
   
\subsubsection{Test 4 (P1): Ensemble Voting System}
Testing for the two proposed ensemble systems (Hierarchical and Best Accuracy) found no discernible difference when tested against the Captivity dataset. This was due to the general high model performance resulting in similar scoring across all models, with the highest scoring combination using the Best Accuracy method, showing the models only scored differently on 0.4\% of the images. 

DenseNet201, MobileNet and ResNet152V2 were found to be the best performing combination of models, so were adopted for the remaining research. Although the high accuracy of the Captivity set did not allow for a meaningful comparison of the approaches, application to the Wild set displayed a more notable pattern. Overall model accuracy using the Hierarchical Approach was found to be 16.7\%, with a \texttt{Wildcat} accuracy of 85.7\%, with Best Accuracy resulting in an overall accuracy of 36.4\% and a \texttt{Wildcat} accuracy of 71.4\%. Whilst the \texttt{Wildcat} accuracy was higher with the Hierarchical approach, this was due to a heavy skew in the data. As such, whilst adopting a Best Accuracy approach does result in a lower accuracy reported for Wildcat in isolation, the higher overall accuracy was viewed to be more beneficial, resulting in a more balanced model. Therefore the Best Accuracy ensemble methods was adopted.

\subsubsection{Test 5 (P1): Segmentation application}
Of the two segmentation methods explored, GrabCut was found to be only partially successful at creating correctly segmented images - either failing completely to segment the image or retaining significant proportions of background data. DeepMAC however was empirically found to perform very well, segmenting animals in the images cleanly in the majority of cases. DeepMAC was therefore adopted for use in subsequent tests were segmentation was applied.

\subsection{Phase 2 Test Results}
Table \ref{tbl:6} details the accuracy results against the Wild test set for all Phase 2 tests undertaken, along with information regarding datasets used to undertake the tests. Accuracies are calculated using the Best Accuracy ensemble method for the three adopted architectures. In addition to the test lists, standard image augmentation approaches such as image flipping and image zoom were progressively added into the models during testing. These were found to have a negligible effect in comparison to the larger changes detailed below.

\subsubsection{Test 6 (P2): Dataset Class Adjustment - Captivity Dataset \textit{Set} and Wildcat Class Separated}
In Test 6a, redistributing the imagery in the \texttt{Wildcat} and \texttt{Animal Unknown} classes appears to have a positive effect on the models' ability to identify wildcat, with Table \ref{tbl:6} showing a 5\% increase from Test 3 in \texttt{Wildcat} accuracy. This improvement was achieved whilst having minimal effect on the overall accuracy and a positive effect on the confusion matrices. The adjustment in dataset was therefore adopted. Test 6b was found to have the opposite effect, with overall \texttt{Wildcat} class dropping by 11.9\% (shown in Table \ref{tbl:6}). Although the model successfully classified some daytime and night-time wildcat imagery into the correct respective class, the separation of the general \texttt{Wildcat} class appears to have weakened the model’s ability to generalise to wildcat overall rather than increasing the likelihood of classification through multiple classes. The separation of the wildcat class was not adopted.

\subsubsection{Test 7: Segmented Image Classification}
Table \ref{tbl:6} shows that despite previous research suggesting that masking the background should aid generalisation, in this case the overall model accuracy actually dropped in comparison to the model trained with the equivalent non-masked dataset (Test 6). However accuracy of the \texttt{Wildcat} class rose. 

Comparison of the relative confusion matrices shows the reduced overall accuracy appears to be a result of a failing of the \texttt{Background} class, although there were also an increased number of images classified as \texttt{Wildcat} and \texttt{Animal Unknown} across all categories. This shows that the segmentation is successful in refocusing what the model considers, but that this refocus may not have been desirable in all contexts. Of particular note is the way the segmentation presents a new hard edge to the classifier (i.e. the edge of the mask), the strength of which may weaken the edge detection layers used to identify the shape of the animal itself. The \texttt{Background} class is also viewed to be a weakness of the models’ construction, with the previously discussed issues around the stronger variability present in the \texttt{Animal Unknown} class.

\subsubsection{Test 8: Two Class Model}
To test the combined impact of the successful experiments, three sub-tests were undertaken. Test 8a considered the segmented dataset used in Test 7, Test 8b uses the equivalent unmasked dataset used in Test 6, and Test 8c uses a new dataset, the dataset from Test 7 but with \texttt{Animal Unknown} imagery removed. Test 8a and Test 8b, by using the same underlying dataset, enabled the performance of segmentation vs no segmentation to be reviewed, alongside testing the two class system itself. Test 8c was a continuation of the research in Test 6, exploring whether removing the model's prior knowledge of more generic wildcat shapes (\texttt{Animal Unknown} being created solely from wildcat movements) could help strengthen the core \texttt{Wildcat} class.

Table \ref{tbl:6} shows that combining the classes resulted in the segmented model (Test 8a) being able to predict \texttt{Wildcat} with 54.8\% accuracy, against an overall model accuracy of 81.6\%. Although the overall accuracy is not comparable with previous results due to the reduction in classes, the \texttt{Wildcat} score represents the highest score achieved on the Wild test set. The unmasked model (Test 8b) comparatively fails to identify \texttt{Wildcat}. Test 8c shows that the removal of \texttt{Animal Unknown} was successful in further increasing the accuracy of the Wildcat class up to 76.2\%, but overall model accuracy suffered and equates to 35.7\% of \texttt{Not Wildcat} imagery being misclassified as \texttt{Wildcat}. Whilst removing 64.3\% of images from consideration is  a positive result for ecologists in the planning / engineering sector, it would need to be associated with an equivalently high threshold of Wildcat detection for it to have beneficial applications.

\subsection{Results Summary}
The tests undertaken in Phase 1 and 2 have systematically experimented with the various components which would be required to create an ensemble classification system and explored a range of techniques currently used in research to encourage model generalisation. The most promising combination found during the research is the use of MegaDetector to identify the location of animals in imagery, DeepMAC to subsequently segment those images, and a trio of image classification models (DenseNet201, MobileNet and ResNet152V2) with two classes (\texttt{Wildcat} and \texttt{Not Wildcat}) to classify those images. An ensembled score, calculated through a Best Accuracy vote, provided the best overall system. This classified \texttt{Wildcat} with 54.8\% accuracy and has an overall model accuracy of 81.6\% against the Wild test set. Table \ref{tbl:6} provides a complete breakdown of all scores.  

\subsection{Environment}

The project was undertaken in two environments, namely the Google Colab cloud environment and a laptop. The laptop environment used TensorFlow 2.5 on a system with an 8 core Intel Xeon CPU with 32GB RAM and an NVIDIA Quadro M1200 GPU chip. Google Colab system specifications vary depending on the system allocated, but for the most part run using an NVIDIA TESLA P100 or NVIDIA V100 GPU with 12GB RAM. TensorFlow models were run in Colab. MegaDetector requires TensorFlow 1.13, DeepMAC requires TensorFlow 2.2 and the renaming models were run on TensorFlow 2.5. Modelling environment requirements are managed as bash commands at the start of notebook scripts, enabling environments to be configured at the time of running. 

\section{Discussion and Conclusion} \label{chap:conc}
The project undertaken has validated previous research in the field that shows whilst it is possible to train a high performing model on camera trap imagery and successfully apply it to an in-sample test set, performance drops are experienced when testing out of sample scenarios. As hypothesised at the outset of this project, the extreme nature of the dataset (using images from a limited number of cameras in a captive enclosure) makes the challenge of generalisation more acute.

The research explored the creation of a system to meet the generalisation challenge by combining pretrained object detection models, an ensemble of classification models which considered images on a local (images cropped to a bounding box) and global (entire camera trap images) scale and image segmentation (along with a range of image manipulations and model class adjustments). Of the eight deep learning model architectures explored, an ensembled combination of DenseNet201, MobileNet and ResNet152V2 were found to give the most accurate prediction, both on a local and global scale. Testing accuracies against the Captivity set were in the region of 90-95\% and 70-75\% for the local and global models respectively. Accuracies against the Wild test set were in the region of 20-45\% and 10-40\%. The significant drop in accuracies led to research being focused on the local models. Experimented adjustments to the structure of the training set and the application of various image manipulation techniques all struggled to significantly alter the Test set accuracy scores.

The most promising results came from applying image segmentation to the local model cropped imagery, removing the background from the images to remove noise. This, combined with focusing the classification models on two classes (\texttt{Wildcat} and \texttt{Not Wildcat}), resulted in an overall test accuracy of 81.6\% and accuracy specific to Wildcat of 54.8\%. Whilst this final performance is an improvement from the original position of the models and provides a direction for future research, it would not in its current state enable an ecologist in planning / engineering to rely upon it for automated classification.

Despite the final accuracy results of the research, the work has proven that the concept of using imagery from captivity to train a model on rare species in the wild is feasible, but that further research is required to fully unlock its potential. This project has shown that the current range of methods used to encourage model generalisation does not seem adequate enough when tested to their limit with a dataset which is relatively numerous in terms of subject examples but not diverse in terms of background context.

A key limitation to the research was the diversity of the imagery used to inform the \texttt{Background} class and the challenges of using iNaturalist data to create an \texttt{Animal Other} class. Suggested future research would be to explore using different datasets to create these classes, creating a more diverse pool of imagery for the \texttt{Background} class and using camera trap imagery and/or UK species to inform the \texttt{Animal Other} class. Other areas for extended research include: investigating whether the masked boundary around a segmented image has a positive or negative effect on classification, the potential to use \textit{style transfer} (Generative Adversarial Networks) to map from the captive imagery data domain to that of the wild (to assist generalisation)\cite{gao_cyclegan-based_2020,karras_style-based_2019}, and the use of monte carlo dropout during testing to estimate confidence of the network \cite{gal_dropout_2016}. 

% This line forces the Acknowledgements to move to the following page - this makes the bottom of the last page line up - which is what they like. If you need more space remove it.
%\vskip 40pt

\section*{Acknowledgment}

With thanks to Zoe Webb and Chris Mellor (both of Ove Arup and Partners) for initially conceiving the project and providing ecological guidance throughout its duration.

\bibliographystyle{IEEEtran}
\bibliography{dissertation_refs}

\end{document}